\documentclass[runningheads]{llncs}
\usepackage{graphicx}
\usepackage[width=122mm,left=12mm,paperwidth=146mm,height=193mm,top=12mm,paperheight=217mm]{geometry}
\usepackage{amsmath,amssymb} 
\usepackage{color}

\usepackage{times}
\usepackage{epsfig}

\usepackage{algorithm}
\usepackage{algorithmic}
\usepackage[table]{xcolor}
\usepackage{multirow}
\usepackage{wrapfig}

\begin{document}
\title{LS-Net: Learning to Solve Nonlinear Least Squares\\ for Monocular Stereo} 

\titlerunning{Learning to Solve NLLS}
%
\author{Ronald Clark\inst{1} \and
Michael Bloesch\inst{1} \and
Jan Czarnowski\inst{1} \and
Stefan Leutenegger\inst{1} \and
Andrew J. Davison\inst{1}
}
%
\authorrunning{Clark, et al.}
%

\institute{Dyson Robotics Lab, Imperial College London, London, SW7 2AZ, UK\\
\email{\{ronald.clark, michael.bloesch, jan.czarnowski, s.leutenegger, a.davison\}@imperial.ac.uk}\\
\url{https://youtu.be/5bZbMm8UqbA}}
\maketitle              
\begin{abstract}
Sum-of-squares objective functions are very popular in computer vision algorithms. However, these objective functions are not always easy to optimize. The underlying assumptions made by solvers are often not satisfied and many problems are inherently ill-posed. In this paper, we propose LS-Net, a neural nonlinear least squares optimization algorithm which learns to effectively optimize these cost functions even in the presence of adversities. Unlike traditional approaches, the proposed solver requires no hand-crafted regularizers or priors as these are implicitly learned from the data. We apply our method to the problem of motion stereo ie. jointly estimating the motion and scene geometry from pairs of images of a monocular sequence. We show that our learned optimizer is able to efficiently and effectively solve this challenging optimization problem.  

\end{abstract}
\section{Introduction}

Most algorithms in computer vision use some form of optimization to obtain a solution that best satisfies some objective function for the problem at hand. The optimization method itself can be seen as simply an intelligent means of searching the solution space for the answer, possibly exploiting the specific structure of the objective function to guide the search.

One particularly interesting form of objective function is one that is composed of a sum of many squared residual terms. 
\begin{equation} E = \frac{1}{2}\sum_j r_j^2(\mathbf{x}) \end{equation}
where $r_j$ is the j-th residual term and $E$ is the optimization objective.

In most cases the residual terms are a nonlinear function of the optimization variables and problems with this type of objective function are called nonlinear least square (NLLS) problems (NLSPs). NLSPs can be efficiently solved using second-order methods \cite{kelley1999iterative}.

However, the success in finding a good solution also depends on the characteristics of the problem itself. The set of residual functions can be likened to a system of equations with their solution at zero, $r_j(\mathbf{x}) = 0$. If the number of variables in this system is larger than the number of equations then the system is underdetermined, if they are equal then it is well-determined and if there are more equations than variables then it is overdetermined. Well-posed problems need to satisfy three conditions: 1) a solution must exist 2) there must be a unique solution and 3) the solution must be continuous as a function of its parameters \cite{Tikhonov:Arsenin:Solutions1977}. 

Undetermined problems are ill-posed as they have infinitely many solutions and therefore no unique solution exists. To cope with this, traditional optimizers use hand-crafted regularizers and priors to make the ill-posed problem well-posed. 

In this paper we aim to utilize strong and well-developed ideas from traditional nonlinear least squares solvers and integrate these with the promising new learning-based approaches. In doing so, we seek to capitalize on the ability of neural network-based methods to learn robust data-driven priors, and a traditional optimization-based approach to obtain refined solutions of high-precision. In particular, we propose to learn how to compute the update based on the current residual and Jacobian (and some extra parameters) to make the NLLS optimization algorithm more efficient and more robust to high noise.

\begin{figure}[b!]
    \centering
    \includegraphics[width=\textwidth]{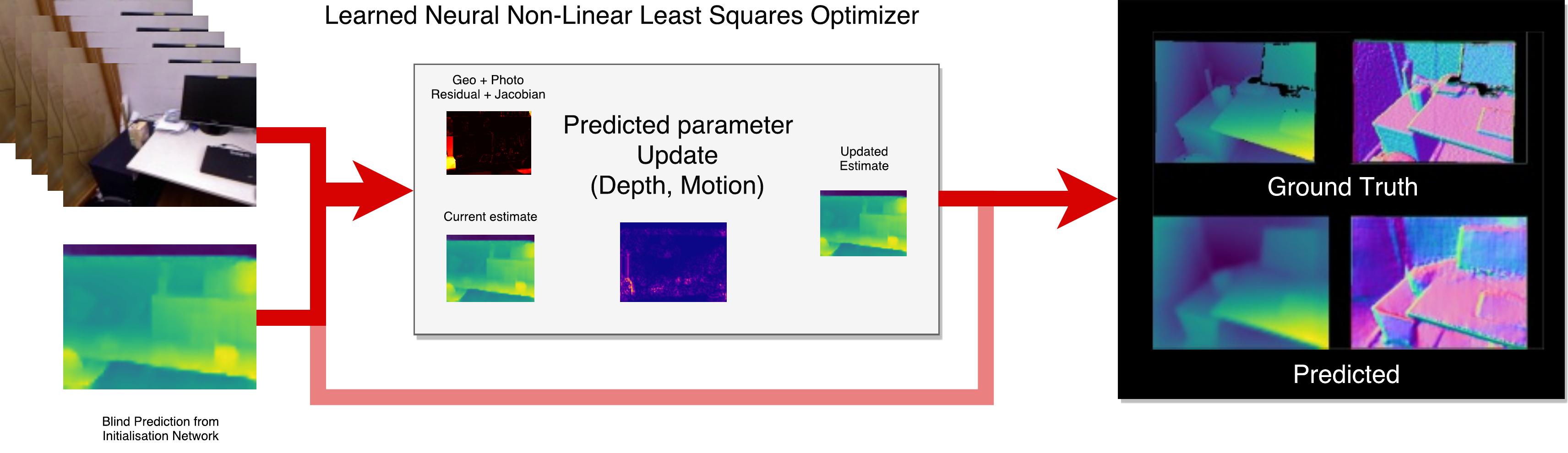}
    \caption{Overview of our system for jointly optimizing a nonlinear least squares objective}
    \label{fig:teaser}
\end{figure}

We apply our optimizer to the problem of estimating the pose and depths of pairs of frames from a monocular image sequence known as monocular stereo as illustrated in Fig. \ref{fig:teaser}.

To summarise, the contributions of our paper are the following:
\begin{enumerate}
    \item We propose an end-to-end trainable optimization method that builds upon the powerful approximate Hessian-based optimization approaches to NLLS problems.
    \item The implicit learning of priors and regularizers for least squares problems directly from data.
    \item The first approach to use a learned optimizer for efficiently minimizing photometric residuals for monocular stereo reconstruction.
\end{enumerate}
Compared to existing learning-based approaches, our method is designed to produce predictions that are accurate and photometrically consistent. 

The rest of the paper is structured as follows. First we outline related work on dense reconstruction using traditional and learning-based approaches. We then visit some preliminaries such as the structure of traditional Gauss-Newton optimizers for nonlinear least square problems. We then introduce our proposed system and finally carry out an evaluation of our method in terms of structure and motion accuracy on a number of sequences from publicly available datasets.

\section{Related Work}

\textbf{Optimization for SLAM}
In visual SLAM we are faced with the problem of estimating both the geometry of the scene and the motion of the camera. This is most often formulated as an optimization over the pixel depths and transformation parameters between pairs of frames. The cost function comprises some form of reprojection error which may be formulated either in terms of geometric or photometric residuals. Geometric residuals require the correspondence of points to be known and thus are only feasible for sparse recostructions. Photometric residuals are formulated in terms of intensity differences and can be computed across the entire image. However, this photometric optimization is difficult as the photometric residuals have high noise levels and various strategies have been proposed to cope with this.
In DTAM \cite{Newcombe:etal:ISMAR2011}, for example, this is accomplished by formulating a cost volume and integrating the residuals from multiple frames before performing the optimization. 
Even then, the residuals need to be combined with a TV-L1 regularization term to ensure noise does not dominate the reconstruction.
Other approaches, such as LSD-SLAM \cite{Engel:etal:ECCV2014}, operate only on high-gradient pixels where the signal-to-noise ratio of the photometric residual is high. 
Even so, none of these systems are able to estimate the geometry and motion in a single joint optimization. Rather, they resort to an approach which swithches between independently optimizing the motion parameters and then the depths in an alternating fashion.
CodeSLAM \cite{Bloesch:etal:CVPR2018} overcomes this problem by using an autoencoder to compress the scene geometry into a small optimizable code, allowing for the joint optimization of both the geometry and motion.

\textbf{Learning for Monocular Stereo}
There has been much interest recently in using end-to-end learning to estimate the motion of a camera \cite{Clark:etal:AAAI2017,Clark:etal:CVPR2017,Wang:etal:ICRA2017} and reconstruct scenes from monocular images \cite{Eigen:etal:ICCV2015}. 
Most of these \cite{Eigen:etal:ICCV2015,Zhou:etal:CVPR2017} are based on feed-forward inference networks.
The training signal for these networks can be obtained in many ways. 
The first approaches were based on a fully-supervised learning signal where labelled depth and pose information were used. 
Subsequent works have shown that the networks can be learned in a self-supervised manner using a learning signal derived, for example, from photometric error of pixel-wise reprojection \cite{Zhou:etal:CVPR2017}, from the consistency of rays projected into a common volume \cite{Tulsiani:CVPR2017} or even using an adversarial signal by modelling the image formation process in a GAN framework \cite{choy20163d}. 
Even so, these approaches only utilize the photometric consistency in an offline manner, i.e.\ during training, and do not attempt to optimize it online as is common in traditional dense reconstruction methods.

To this extent, some works such as \cite{Ummenhofer:Brox:CVPR2017}, have demonstrated that it is beneficial to include multiple views and a recurrent refinement procedure in the reconstruction process. Their network, comprising three stages, is closely related to the structure which we build on in this work. 
The first stage consists of a \textit{bootstrap network} which produces a rough low-resolution prediction; the second stage consists of an \textit{iterative network} which iteratively refines the bootstrap prediction; and finally a \textit{refinement network} which computes a refined and upscaled depth map. 

In this paper, we adopt the same structure but formalize the iterative network as an optimization designed to enforce multiview photometric consistency where the bootstrap network acts as an initialization of the optimization and the refinement acts as an upscaling.
In essence, our reconstruction is based on an optimization procedure that is itself optimized using data. This is commonly referred to in the machine-learning literature as a \textit{meta-learned optimizer}.

\textbf{Meta-learning and Learning to Optimize} A popular and very promising avenue of research which has been receiving increasing attention is that of meta-learned optimizers. 
Such approaches have shown great utility in performing few-shot learning without overfitting \cite{Ravi:ICLR2016,clark:etal:metalearn}, for optimizing GANS which are traditionally very difficult to train \cite{Metz:2016}, for optimizing general black box functions \cite{Chen:ICML2017} and even for solving difficult combinatorial problems \cite{Dai:2017}. 
Perhaps the most important advantage is to learn data-driven regularization as demonstrated in \cite{Oktem:IOP2017} where the authors use a partially learned optimization approach for solving ill-posed inverse problems. In \cite{lin2016inverse}, the authors train through a multi-step inverse compositional Lukas Kanade algorithm for aligning 2D images. 
In our method, we utilize a learned multi-step optimization model by using a recurrent network to compute the update steps for the optimization variables. While most approaches that attempt to learn optimization updates use either fully learned update steps \cite{Xiong:etal:CVPR2013,Li:ICLR2017} about the objective and first-order gradient information \cite{Chen:ICML2017}, we exploit the least-square structure of our problem and forward the full Jacobian matrix to provide the network with richer information. Our approach is -- to the best of our knowledge -- the first to use second-order approximations of the objective to learn optimization updates.

\section{Preliminaries}

\subsection{Nonlinear Least Squares Solvers}
Many optimization problems have an objective that takes the form of a sum of squared residual terms, $E = \frac{1}{2}\sum_j r_j^2(\mathbf{x})$ where $r_j$ is the j-th residual term and $E$ is the optimization objective. As such, much research has been devoted to finding efficient solvers for problems of this form. Two of the most successful and widely used approaches are the Gauss-Newton (GN) and Levenberg-Marquadt (LM) methods. Both of these are second-order, iterative optimization methods. However, instead of computing the true Hessian, they exploit the least-squares structure of the objective to compute an approximate Hessian that is used in the updates. Given an initial estimate of the variables, $\mathbf{x}_0$, these approaches compute updates to the optimization variable in the attempt to find a better solution, $\mathbf{x}_i$, at each step $i$. The incremental update, $\Delta \mathbf{x}_i$ is computed by solving a linear least squares problem which is formed by linearising the residual at the current estimate $\mathbf{r}(\mathbf{x}_i + \Delta \mathbf{x}_i) \approx \mathbf{r}_i + \mathbf{J}_i \Delta \mathbf{x}_i$ \cite{kelley1999iterative}, with the abbreviations:
\begin{equation}
    \mathbf{r}_i = \mathbf{r}(\mathbf{x}_i), \ \ \ \ \mathbf{J}_i = \left.\frac{d\mathbf{r}}{d\mathbf{x}} \right\vert_{\mathbf{x} = \mathbf{x}_i}.
\end{equation}

Using the linearized residual, the optimal update can be found as the solution to the quadratic problem \cite{kelley1999iterative}
\begin{equation}
\Delta \mathbf{x}_i = \underset{ \Delta \mathbf{x}_i}{\arg\min} \frac{1}{2}|| \mathbf{r}_i + \mathbf{J}_i \Delta \mathbf{x}_i ||^2.
\end{equation}
The well known Normal equations to this can be computed analytically by differentiating the problem and equating to zero. The update step used in GN is then given by solving:
\begin{equation}
      \mathbf{J}_i^T \mathbf{J}_i \Delta \mathbf{x}_i = - \mathbf{J}_i^T \mathbf{r}_i
\end{equation}
By comparing this to Newton's method which requires the computation of the true Hessian $\mathbf{H}(\mathbf{x}_i)$ for finding updates \cite{fletcher2013practical}, we see that the GN method effectively approximates $\mathbf{H}(\mathbf{x}_i)$ using $ \mathbf{J}_i^T \mathbf{J}_i$, which is usually more efficient to compute. LM extends GN by adding a damping factor $\lambda$ to the update $\Delta \mathbf{x}_i = - (\mathbf{J}_i^T \mathbf{J}_i+\lambda \, \mathrm{diag}(\mathbf{J}_i^T \mathbf{J}_i))^{-1} \mathbf{J}_i^T \mathbf{r}_i$ to better condition the updates and make the optimization more robust \cite{fletcher2013practical}.

In our proposed approach, we build on the GN method by not restricting the updates to be a static function of $\mathbf{J}_i$. Compared to LM which adaptively sets a single parameter, $\lambda$, we compute the entire update step by using a neural network which has as its input the full Jacobian $\mathbf{J}_i$. The details of this are described in Section \ref{sec:optimization}.

\subsection{Warping and Photometric Cost Function}

The warping function we use for the least squares cost function is similar to the loss used in the usupervised training in \cite{Zhou:etal:CVPR2017}. The warping is based on a spatial transformer which first transforms the coordinates of points in the target view to points in the source view and then samples the source view. The 4x4 transformation matrix, $\hat{T}_{t\rightarrow s}$ is obtained by applying an exponential map to the output of the network, i.e.  $\hat{T}_{t\rightarrow s}= \exp{(\mathbf{p}^\times)}$ where $\mathbf{p}$ (bold face) is the relative pose represented as a six-vector and $p_s$ (non-bold face) is the pixel location in the source image and $p_t$ (non-bold face) is a pixel location in the target image (consistent with the notation in the paper)

\begin{equation}
    p_s \sim K \hat{T}_{t\rightarrow s}\hat{D}_{t}(p_t)K^{-1}p_t
    \label{eq:projection}
\end{equation}

Using these warped coordinates, a synthesized image $\hat{I}_s(p)$ is obtained through bilinear sampling of the the source view at the locations $p_s$ computed in Eqn. \ref{eq:projection}. The least squares loss function from which we derive $\mathbf{J}$ is then, 

\begin{equation}
    L = \sum_p || I_t (p) - \hat{I}_s(p) ||_2~,
    \label{eq:objective}
\end{equation}

where $I_t$ and $I_s$ are the source and target intensity images and the residual corresponding to each pixel is $\mathbf{r}_p = I_t (p) - \hat{I}_s(p)$. The elements of the Jcaobian of the warping function, $\mathbf{J}$, can be easily computed using autodiff (in Tensorflow simply {\tt tf.gradients(res[i],x))} for each residual. However, to speed up our implementation we anylytically compute the elements of the Jacobian in our computation graph.

\section{Model}
The model is built around the optimization of the photometric consistency of the depth and motion predictions for a short sequence of input images. 
Each sequence of images has a single ``target" keyframe (which we choose as the first frame) for which we optimize the depth values. In all cases, we operate on inverse depths, $z = \frac{1}{d}$ for better handling of large depths values.
Our model additionally seeks to optimize for the relative transformations between each source frame $s$ in the sequence and the target keyframe $t$, $\mathbf{p}_{t\rightarrow s}$.
The full model consists of three stages. 
All iterative optimization procedures require an initial starting point and thus the initialization stage serves the purpose of predicting a good initial estimate. 
The optimization stage consists of a learned optimizer which benefits from explicitly computed residuals and Jacobians. 
To make the optimization computationally tractable, the optimization network operates on a down-sampled version of the input and exploits the sparsity of the problem. 
The final stage of the network upsamples the prediction to the original resolution. 
The networks (including those of the optimizer) are trained using a supervised loss.
We now describe each of the three network components in detail.

\begin{algorithm}
\caption{Opitimizing least squares with LS-Net}
\label{alg:opt_alg_structure}
\begin{algorithmic}
\REQUIRE Residual function $\mathbf{r}(\mathbf{x})$, image sequence $\mathbf I_1,\mathbf I_2, \ldots $
\STATE $\mathbf{x}_{0} \gets  f_{\theta_0} (\mathbf I_1,\mathbf I_2,\ldots )$
\FOR{$i = 0,1,\ldots N-1$}
    \STATE $\Delta \mathbf{x}_i, \mathbf{h}_{i+1} \gets f_\theta \left(\Phi(\mathbf{J}_i, \mathbf{r}_i), \mathbf{h}_{i} \right)$
    \IF{$||\Delta \mathbf{x}_i|| < \epsilon$ }
        \RETURN{$\mathbf{x}_{i}$}
    \ENDIF
    \STATE $\mathbf{x}_{i+1} \gets \mathbf{x}_{i} + \Delta \mathbf{x}_i$
\ENDFOR
\end{algorithmic}
\end{algorithm}

\subsection{Initialization Network}

The purpose of the initialization network is to predict a suitable starting point for the optimization stage.
We provide the initialization network with both RGB images and thereby allow it to leverage stereopsis.
The architecture of this stage is a simple convolutional network. 
For this stage we use 3 convolutions with stride 2, one convolution with stride 1 and one upsamplings + convolutional layers. This results in the output of the network being downscaled by a factor of 4 for feeding into the optimization stage. The network also produces an initial pose using a fully connected layer branched from the central layers of the network.
Thus the output of the initialization stage consists of an initial depth image and pose.

\subsection{Optimizing Nonlinear Least Squares with LS-Net}

\label{sec:optimization}

The learnt optimization procedure is outlined in Algorithm \ref{alg:opt_alg_structure}. The optimization network attempts to optimize the photometric objective $E(\mathbf{x})$ where $\mathbf{x} = (\mathbf{z},\mathbf{p})$ are the optimization variables (inverse depths $\mathbf{z}$ and pose $\mathbf{p}$). The objective $E(\mathbf{x})$ is a nonlinear least squares expression defined in terms of the photometric residual vector $\mathbf{r}(\mathbf{x})$
\begin{align}
    E(\mathbf{x}) = \frac{1}{2} ||\mathbf{r}(\mathbf{x})||^2.
\end{align}

The updates of the parameters to be optimized, $\mathbf{x}$, follow a standard iterative optimization scheme, i.e.
\begin{equation}
{\mathbf{x}}_{i+1} = {\mathbf{x}}_i + \Delta \mathbf{x}_i.
\label{eqn:update}
\end{equation}
In our case, the updates $\Delta \mathbf{x}_i$ are predicted using a Long Short Term Memory Recurrent Neural Network (LSTM-RNN) \cite{Hochreiter:2001}. 
In order to compute the Jacobian we use automatic differentiation available in the Tensorflow library \cite{abadi2016tensorflow}. 
Using the automatic differentiation operation, we add operations to the Tensorflow computation graph \cite{abadi2016tensorflow} which compute the Jacobian of our residual vector with respect to the dense depth and motion. 
As the structure of the Jacobian often exhibits problem specific properties, we apply a transformation to the Jacobian, $\Phi(\mathbf{J}_i, \mathbf{r}_i)$ before feeding this Jacobian into our network. The operation $\Phi$ may involve element-wise matrix operations such as gather or other operations which simplify the Jacobian input. The operations we use for the problems addressed in this paper are detailed in Section \ref{sec:sparsity}.

To allow for the computation of parameter updates which are not restricted to those derived from the approximate Hessian, we turn to the powerful function approximation ability of the LSTM-RNN \cite{Hochreiter:2001} to learn the final parameter update operation from data. As the number of coordinates are likely to be very large for most optimization problems, \cite{Chen:ICML2017} propose to use one LSTM-RNN for each coordinate. For our problem, we have Jacobians with high spatial correlations and thus we replace the coordinate-wise LSTM with a convolutional LSTM. The per-iteration updates, $\Delta \mathbf{x}_i$ are predicted by a network which in this case is an LSTM-RNN,
\begin{equation}
        \begin{bmatrix}
            \Delta x_i  \\ h_{i+1}
        \end{bmatrix}
        = \mbox{LSTM}_{cell}\left(\Phi(\mathbf{J}_i, \mathbf{r}_i), h_i, \mathbf{x}_i; \mathbf{\theta}\right),
\end{equation}
where $\mathbf{\theta}$ are the parameters of the networks and $\mbox{LSTM}_{cell}$ is a standard LSTM cell update function with hidden layer $h_i$. 


\subsection{The Jacobian input structure}
\label{sec:sparsity}

\begin{figure}
    \centering
    \includegraphics[width=\columnwidth]{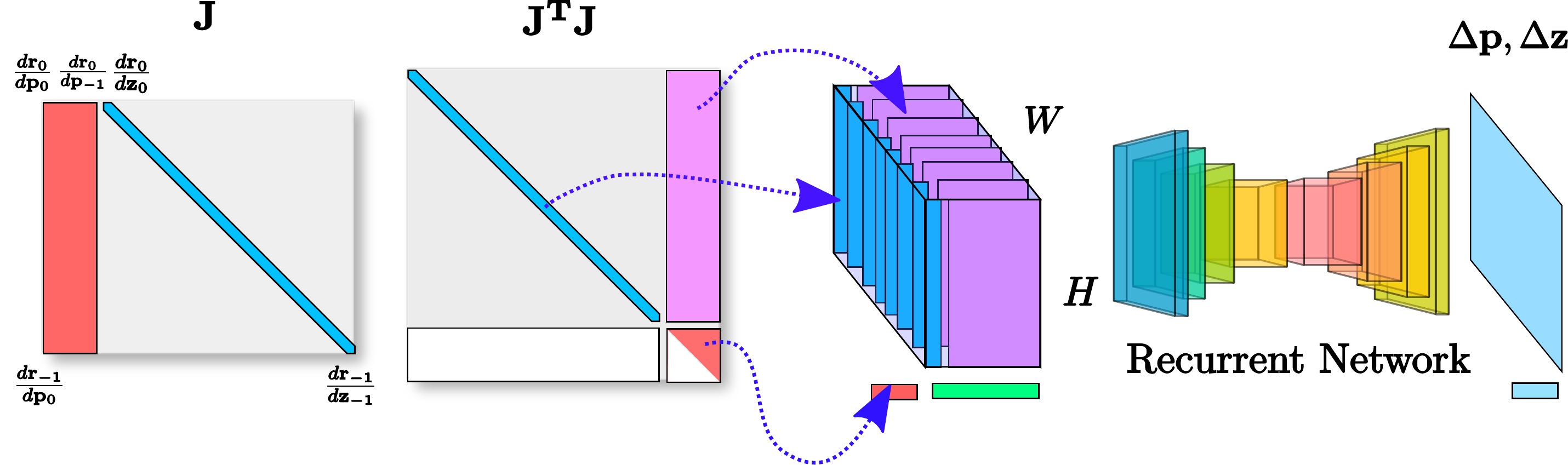}
    \caption{The block-sparsity structure of $\mathbf{J}$ and $\mathbf{J}^T\mathbf{J}$ for the depth and egomotion estimation problem.}
    \label{fig:sparsity}
\end{figure}

Each type of least squares cost function gives rise to a special Jacobian structure. The input function, $\Phi(\mathbf{J}, \mathbf{r})$, to our network serves two purposes; one functional and the other structural. Firstly, $\Phi$ serves to compute the \emph{approximate Hessian} as is done with the classical Gauss-Newton optimization method:
\begin{equation}
\Phi(\mathbf{J}, \mathbf{r}) = [\mathbf{J^T} \mathbf{J},\mathbf{r}].
\end{equation}
The structure of $\Phi(\mathbf{J}, \mathbf{r})$ is shown in Figure \ref{fig:sparsity}. We note that we choose not to compute the full $(\mathbf{J}^T \mathbf{J})^{-1}\mathbf{J}$ as this adds additional computational complexity to the operation which is repeated many times during training.
We also compress the sparse $\mathbf{J}^T \mathbf{J}$ into a compact form as illustrated in Figure \ref{fig:sparsity}.
The output of this restructuring yields the same image shape as the image. The compressed structure allows efficient processing of the matrix.

\subsection{Upscaling Network}
As the optimization network operates on low-resolution predictions, an upscaling network is used to produce outputs of the desired size. The upscaling network consists of a series of bilinear upsampling layers concatenated with convolutions and acts as a super-resolution network. The input to the upscaling network consists of the low-resolution depth prediction and the RGB image.

\section{Loss Function}
\label{sec:losses}

In this section we describe the loss function which we use to train the network weights of all three stages of our model.

The current state-of-the-art depth and motion prediction networks still rely on labelled images to provide a strong learning signal. 
We include a loss term based on labelled ground truth inverse depth images $\tilde{\mathbf{z}}$,  

\begin{equation}
L_{depth}(\mathbf{x}) = \frac{1}{wh} \| \mathbf{z} - \tilde{\mathbf{z}} \|_1
\end{equation}
with image width $w$ and height $h$, and where $\mathbf{z}$ is the predicted inverse depth image.

We also use a loss term based on the relative pose between the source (s) and target (t) frame, $\tilde{\mathbf{p}} = (\tilde{\mathbf{t}}_{t\rightarrow s}, \tilde{\boldsymbol{\alpha}}_{t\rightarrow s}) $ with translation $\tilde{\mathbf{t}}_{t\rightarrow s}$ and rotation vector $\tilde{\boldsymbol{\alpha}}_{t\rightarrow s}$ from ground-truth data,
\begin{equation}
L_{pose}(\mathbf{x}) =  \sum_{s} \| \boldsymbol{\alpha}_{t\rightarrow s} - \tilde{\boldsymbol{\alpha}}_{t\rightarrow s}\|_1 +  \| \mathbf{t}_{t\rightarrow s} - \tilde{\mathbf{t}}_{t\rightarrow s}\|_1
\end{equation}
Note that this loss function need not be a sum of squares and can be computed using any other form using eg. L1 etc. The final loss function consists of a weighted combination of the individual loss terms:

\begin{align}
    L_{tot}(\mathbf{\theta}) = \sum_{i} w_{pose} L_{pose}(\mathbf{x}_i(\mathbf{\theta}))+ w_{depth} L_{depth}(\mathbf{x}_i(\mathbf{\theta})). \label{eq:ltot}
\end{align}
Note that our objective here includes the ground-truth inverse depth which we do not have access to when computing the residuals $\mathbf{r}$ (and then the Jacobian $\mathbf{J}$) in the recurrent optimization network in Section \ref{sec:optimization}.

The optimization network is never directly privy to the ground truth depth and poses, it only benefits from these by what is learned in the network parameters during training.
In this manner, we have a system which is trained offline to best minimize our objective online. 
During the offline training phase, our system learns robust priors for the optimization by using the large amounts of labelled data. 
During the online phase our system optimizes for photometric consistency only but is able to utilize the knowledge it has learned during the offline training to better condition the optimization process.

\section{Training}

During the training, we unroll our iterative optimization network for a set number of steps and backpropogate the loss through the network weights, $\theta$. 
In order to find the parameters of the optimizer network, the meta-loss, $L_{tot}(\mathbf{\theta})$, is minimized using the ADAM optimizer where the total meta-loss is computed as the loss summed over the $N$ iterations of the learned optimization (see Eq. \ref{eq:ltot}). 
For each step $i$ in the optimization process we update the state $\mathbf{x}_i$ of the optimization network according to Eqn. \ref{eqn:update}.

As our loss depends on variables which are updated recurrently over a number of timesteps, we use backpropogation through time to train the network. 
Backpropogation through time unrolls each step and updates the parameters by computing the gradients through the unrolled network.
In our experiments we unroll our optimization for 15 steps.

We find that training the whole network at once is difficult and thus train the initialization network first before adding the optimization stage.

\section{Evaluation}

In this section we evaluate the proposed method on both synthetic and real datasets. We aim to determine the efficiency of our approach i.e. how quickly it converges to an optimum and how it compares to a network which does not explicitly incorporate the problem structure in its iterations.

\subsection{Synthetic data experiments}
In this section we evaluate the performance of our proposed method on a number of least squares curve fitting problems. We experiment on curves parameterized by two variables, $\mathbf{x} = (a,b)$. We chose a set of four functions to use for our experiment as follows
\begin{equation}
    y = x \exp(a t) + x\exp(b t) + \epsilon,
    \label{eqn:eqn1}
\end{equation}
\begin{equation}
    y = \sin(a t + b) + \epsilon,
\end{equation}
\begin{equation}
    y = \mathrm{sinc}(a t + b) + \epsilon,
\end{equation}
\begin{equation}
    y = \mathcal{N}(t|\mu=a,\sigma=b) \hspace{5pt} \mbox{(fitting a Gaussian)}
    \label{eqn:eqn4}
\end{equation}

\begin{figure}
\centering
   \includegraphics[width=0.32\linewidth]{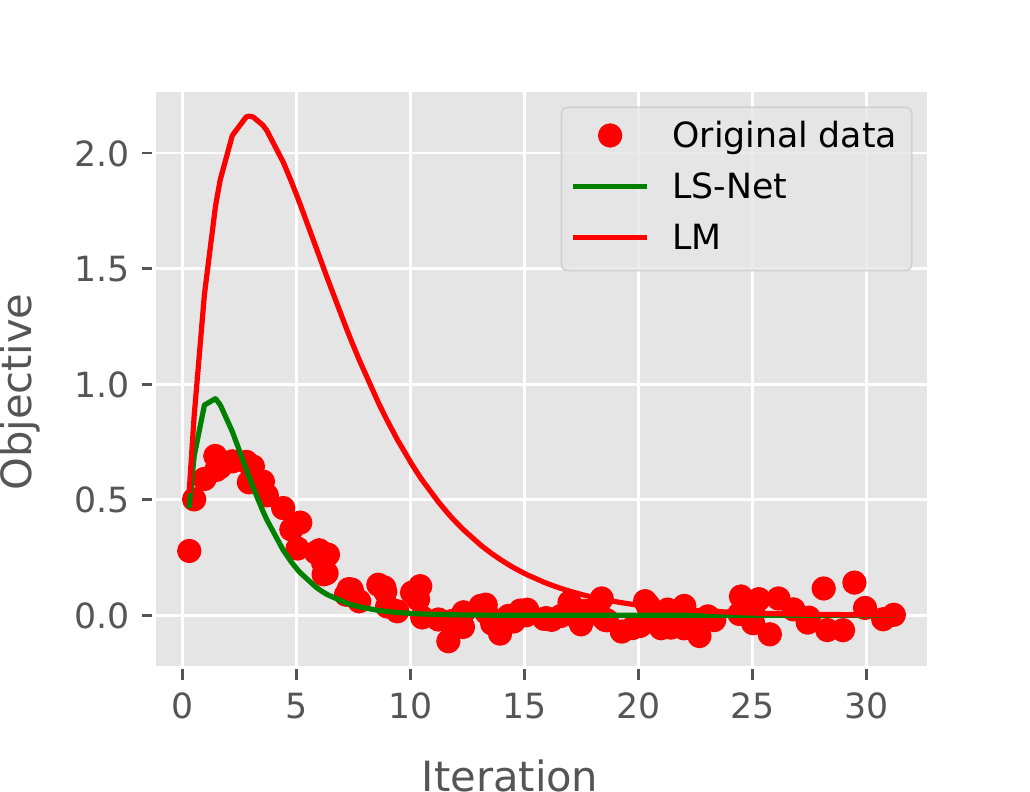}
   \includegraphics[width=0.34\linewidth]{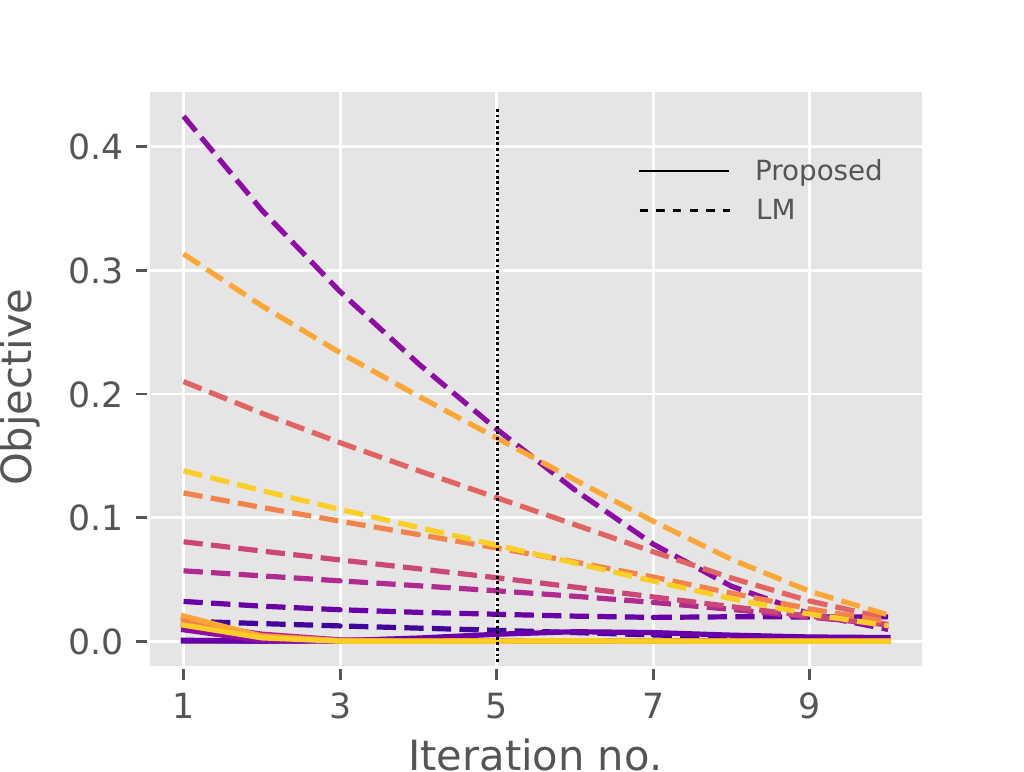}
   \includegraphics[width=0.30\linewidth]{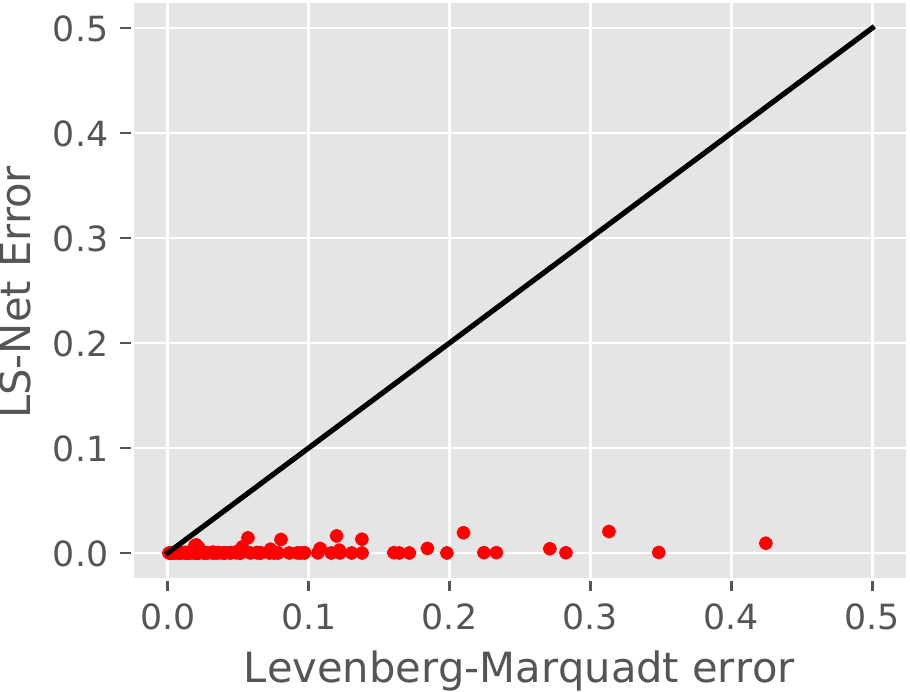}
\caption{Comparison between our method and standard least squares for fitting parametric functions to noisy data with a least-squares objective. In a) the fitted functions limited to 5 iterations is shown, in b) the error as a function of iteration no. is shown for 10 test functions and in c) the LM error is plotted against the error of the proposed method for all iterations.}
\label{fig:syn_results}
\end{figure}

For these experiments we generate the data by randomly sampling one of four parametric functions (Eqn.\ \ref{eqn:eqn1} to Eqn.\ \ref{eqn:eqn4}) as well as the two parameters $a$ and $b$. For the training data we add noise $\epsilon \sim \mathcal{N}(0,0.1)$ to the true function values. In Figure \ref{fig:syn_results} we show the results on a test set of sampled functions. Figure \ref{fig:syn_results} a) shows the fitted function after 5 iterations (of a total of 15 iterations) for our method and standard LM. The learned approach clearly outperforms LM in terms of speed of convergence. In Figure \ref{fig:syn_results} b) we see the learned errors vs LM for all steps in the optimization, where again, the learned method clearly outperforms LM.

\subsection{Real-world test: depth and pose estimation}
In this section we test the ability of our proposed method on estimating the depth and egomotion of a moving camera.
To provide a fair evaluation of the proposed approach, we use the same evaluation procedure as in \cite{Ummenhofer:Brox:CVPR2017} and report the same baselines, where oracle uses MVS with known poses, \textit{SIFT} uses sparse-feature for correspondences, \textit{FF} uses optical flow, \textit{Matlab} uses the KLT tracker in Matlab as the basis of a bundle-adjusted reconstruction.

\begin{figure*}[h!]
    \centering
    \includegraphics[width=\textwidth]{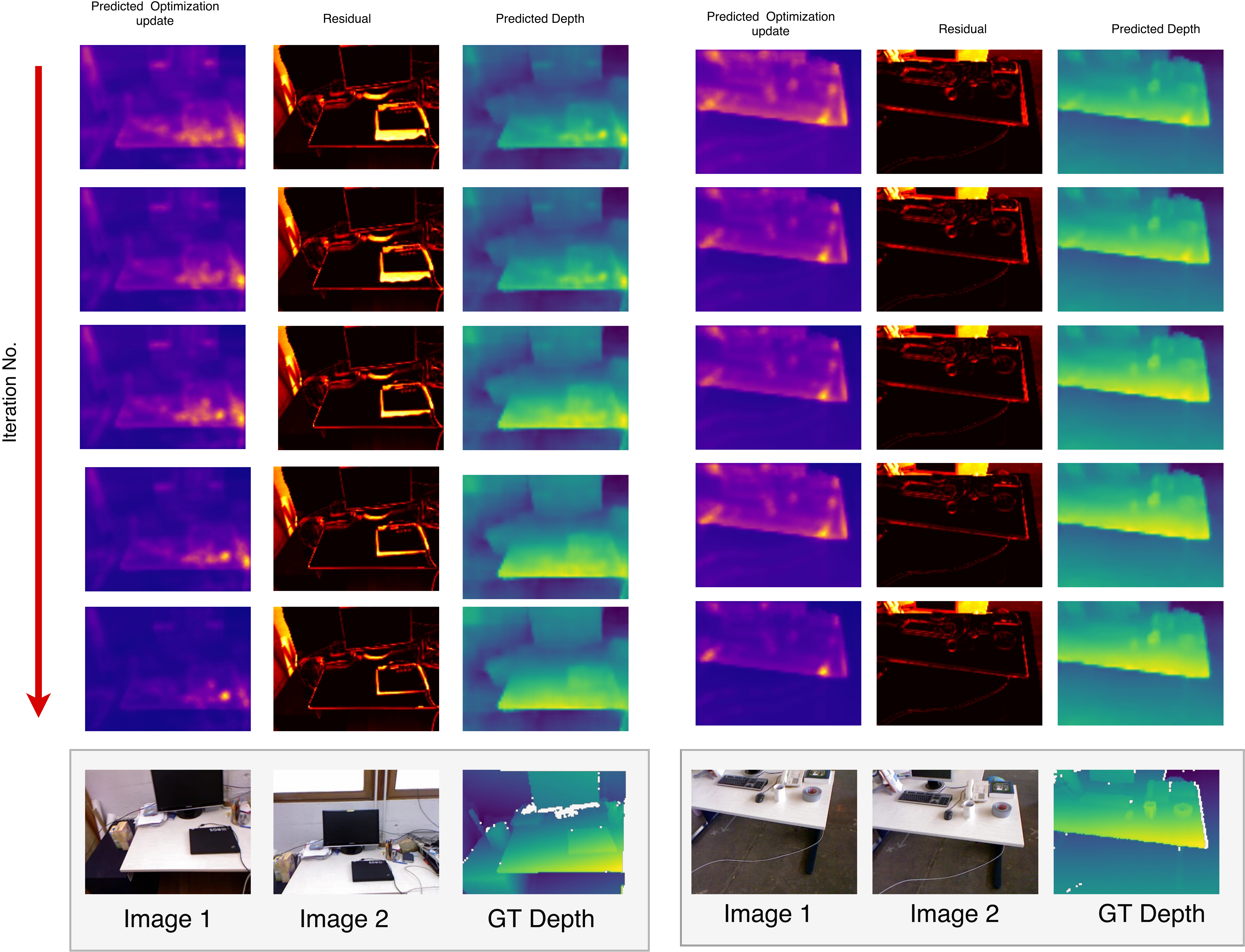}
    \caption{Qualitative results on two challenging indoor scenes using only two frames. The figure shows the last 5 iterations of 15 of the optimization network. Even with this wide baseline, and only two frames, our method is able to optimize the photometric error reliably. }
    \label{fig:results}
\end{figure*}

\subsection{Metrics}

We evaluate the performance of our approach on the depth as well as the motion prediction performance. 
For depth prediction we use the absolute, scale-invariant and relative performance metrics. 

\subsection{Datasets}

The datasets which we use to evaluate the network consist of both indoor and outdoor scenes. For all the datasets, the camera undergoes free 6-DoF motion. To train our network we use images from all the datasets partitioned into testing and training sets.

\noindent \textbf{MVS} The multiview stereo dataset consists of a collection of scenes obtained using struction from motion software followed by dense multi-view stereo reconstruction. We use the same training/test split as in \cite{Ummenhofer:Brox:CVPR2017}. The training set of images used included ``Citywall", ``Achteckturm" and ``Breisach" scenes with ``Person-Hall", ``Graham-Hall", and ``South-Building" for testing.

\noindent \textbf{TUM} The TUM RGB-D dataset consists of Kinect-captured RGB-D image sequences with ground truth poses obtained from a Vicon system. It comprises a total of 19 sequences with 45356 images. We use the same test / train split as in \cite{Ummenhofer:Brox:CVPR2017} with 80 held-out images for test.

\noindent \textbf{Sun3D} The SUN3D dataset consists of scenes reconstructed using RGB-D structure-from-motion. The dataset has a variety of indoor scenes, with absolute scale and consists of 10,000 individual images. The poses are less accurate than the TUM dataset as they were obtained using an RGB-D reconstruction.

\begin{figure}[h!]
    \centering
    \includegraphics[width=1.0\columnwidth]{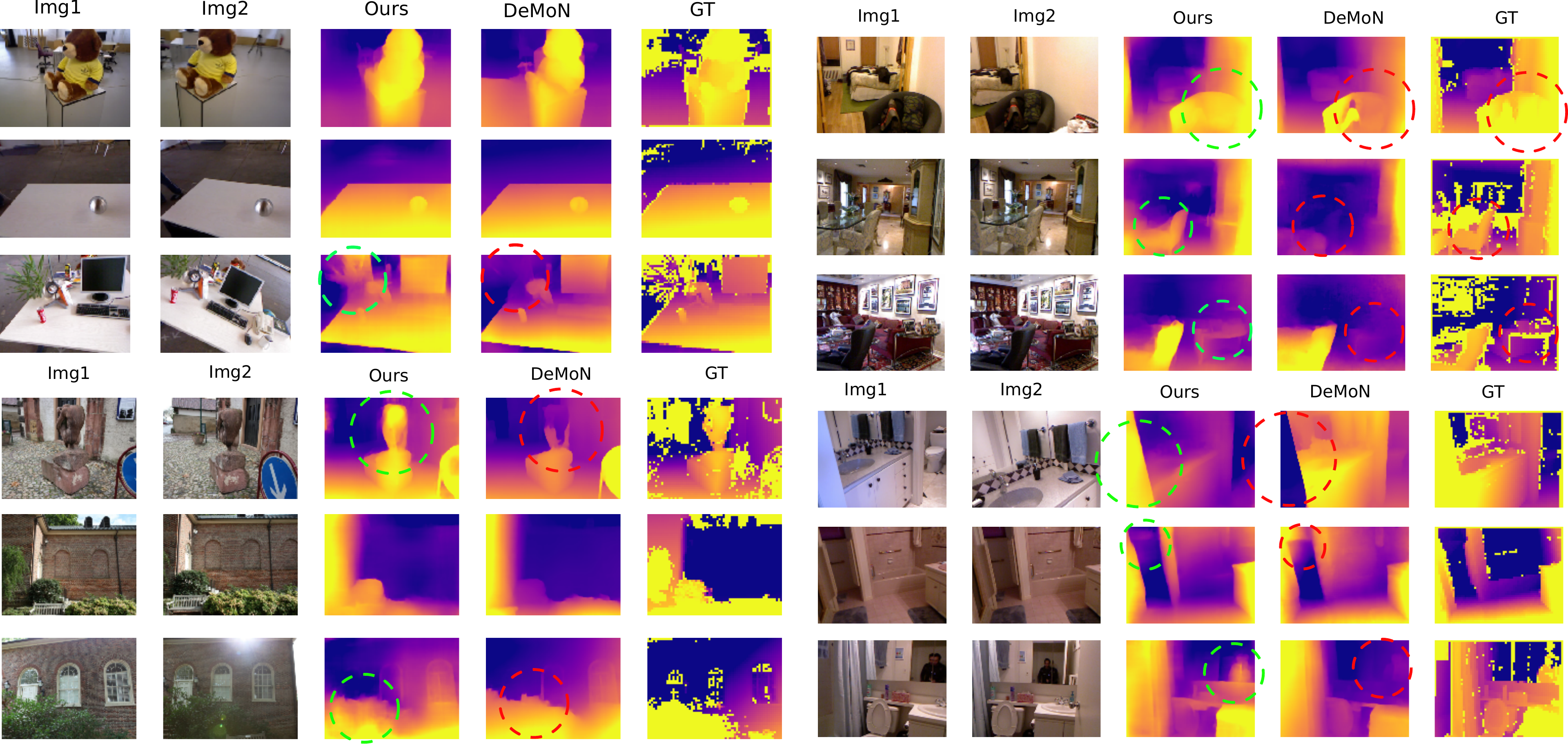}
    \caption{Qualitative results on the NYU dataset. Compared to DeMoN our network has fewer "hallucinations" of structures which do not exist in the scene. }
    \label{fig:qualitative_comparison}
\end{figure}


A qualitative evaluation of our method compared to standard multiview stereo and DeMoN \cite{Ummenhofer:Brox:CVPR2017} is shown in Figure \ref{fig:qualitative_comparison}. Our method produces depth maps with sharper structures compared to DeMoN, even with a lower output resolution.  Compared to COLMAP \cite{schoenberger2016mvs} our reconstruction is more dense and does not include as many outlier pixels. Numerical results on the testing data-sets are shown in Table \ref{tbl:results1}. As is evident from the Table, our learned optimization approach outperforms most of the traditional baseline approaches, and performs better or on par with DeMoN on most cases. This may be due to our architectural choice as we do not include any alternating flow and depth predictions.

\begin{table}[h!]

\centering
  \begin{tabular}{c|c|ccc|cc}
      \hline

  &             & \multicolumn{3}{c|}{Depth}             & \multicolumn{2}{c}{Motion}    \\
      \hline
& Method      &  L1-inv      &  sc-inv    &  L1-rel    & Rotation        & Translation             \\
      \hline
      \hline
      \parbox[t]{4mm}{\multirow{6}{*}{\rotatebox[origin=c]{90}{MVS}}}      & MVS & 0.019   & 0.197 & \cellcolor{green!25}0.105 &  0         &  0                \\
& SIFT   &   0.056      &   0.309    &   0.361    &  21.180    &  60.516           \\
& FF     &   0.055      &   0.308    &   0.322    & 4.834 &  17.252           \\
& Matlab &     -        &     -      &     -      &  10.843    &  32.736           \\
& DeMoN       &   \cellcolor{green!25} 0.047      &   0.202    &   0.305    &   5.156    &  14.447      \\
& Proposed       &   0.051     &  \cellcolor{green!25}0.221     &      0.311        &    \cellcolor{green!25}4.653         &     \cellcolor{green!25} 11.221 \\
      \hline
      \hline
      \parbox[t]{2mm}{\multirow{6}{*}{\rotatebox[origin=c]{90}{Scenes11}}} & Oracle &   0.023      &   0.618    &   0.349    &  0         &  0                \\
& SIFT   &   0.051      &   0.900    &   1.027    &  6.179     &  56.650           \\
& FF     &   0.038      &   0.793    &   0.776    &  1.309     &  19.425           \\
& Matlab &     -        &     -      &     -      &  0.917     &  14.639           \\
& DeMoN       &   0.019      & \cellcolor{green!25}0.315       & 0.248 & \cellcolor{green!25}0.809 &  8.918       \\
& Proposed       &  \cellcolor{green!25}0.010    & 0.410      & \cellcolor{green!25} 0.210    &  0.910     & \cellcolor{green!25} 8.21       \\
      \hline
      \hline
      \parbox[t]{2mm}{\multirow{6}{*}{\rotatebox[origin=c]{90}{RGB-D}}}    & Oracle &  0.026  &   0.398    &   0.336    &  0         &  0                \\
& SIFT   &   0.050      &   0.577    &   0.703    & 12.010     &  56.021           \\
& FF     &   0.045      &   0.548    &   0.613    &  4.709     &  46.058           \\
& Matlab &      -       &     -      &     -      &  12.831    &  49.612
\\
& DeMoN       &   0.028      & 0.130 & 0.212 & 2.641 &  \cellcolor{green!25}20.585      \\
& Proposed       & \cellcolor{green!25}0.019       & \cellcolor{green!25}0.09   & 0.301     &  \cellcolor{green!25}1.01    &  22.1    \\
      \hline
      \hline
      \parbox[t]{2mm}{\multirow{6}{*}{\rotatebox[origin=c]{90}{Sun3D}}}    & oracle &   0.020      &   0.241    &   0.220    & 0          &  0                \\
& SIFT   &   0.029      &   0.290    &   0.286    & 7.702      &  41.825           \\
& FF     &   0.029      &   0.284    &   0.297    & 3.681      &  33.301           \\
& Matlab &      -       &     -      &     -      & 5.920      &  32.298
\\
& DeMoN       &   0.019 &\cellcolor{green!25} 0.114 & \cellcolor{green!25}0.172 & 1.801 &  18.811      \\
& Proposed       & \cellcolor{green!25} 0.015 & 0.189   &  0.650    &  \cellcolor{green!25}1.521    &  \cellcolor{green!25}14.347      \\
      \hline

    \end{tabular}
    \vspace{5mm}
    \caption{Quantitative results on the evaluation datasets. Green highlights the best performing method for a particular task.}
    \label{tbl:results1}
\end{table}

\section{Discussion}
In the context of optimisation, our network-based updates accomplish something which a classical optimisation approach cannot in that it is able to reliably optimise a large under-determined system with implicitly learned priors. 
\begin{wraptable}{l}{0.5\textwidth}
\label{tbl:comparison}
\vspace{-25pt}
\caption{Summary of the performance of LS-Net optimisation compared to standard LM. Table indicates the best performing method for criteria.}
\centering
\begin{tabular}{c|ccc} \hline
& \multicolumn{3}{c}{ Problem Size } \\ \hline
& Small & Medium & Large \\ \hline
Accuracy & \cellcolor{green!25} Ours & \cellcolor{green!25} Ours &  \cellcolor{green!25}Ours \\
Memory & Tie & \cellcolor{green!25}Ours & \cellcolor{green!25}Ours \\
Speed &  Tie & \cellcolor{green!25}Ours & \cellcolor{green!25}Ours \\ \hline
\end{tabular}
\vspace{-18pt}
\end{wraptable}
 For a large under-determined problem like in the depth and motion case, standard Levenberg-Marquadt (LM) fails to improve the objective and the required sparse matrix inversion for a $J^TJ$ with $\approx91K$ non-zero elements (128 $\times$ 96 size image) takes 532ms, compared to our network forward pass which takes 25ms. For small, overdetermined problems LM does work and for this reason, in Section 7.1, we have compared our approach to LM on a small curve fitting problem and found that our approach significantly outperforms it in terms of accuracy and convergence rate. For the small problem, the matrix inversion in the standard approach (LM) is very quick but we are also able to use a smaller network so our time per-iteration is tie with LM.  This is summarised in Table \ref{tbl:comparison}.

\subsection{Ablation study}

\begin{figure}
    \centering
    \includegraphics[width=0.5\columnwidth]{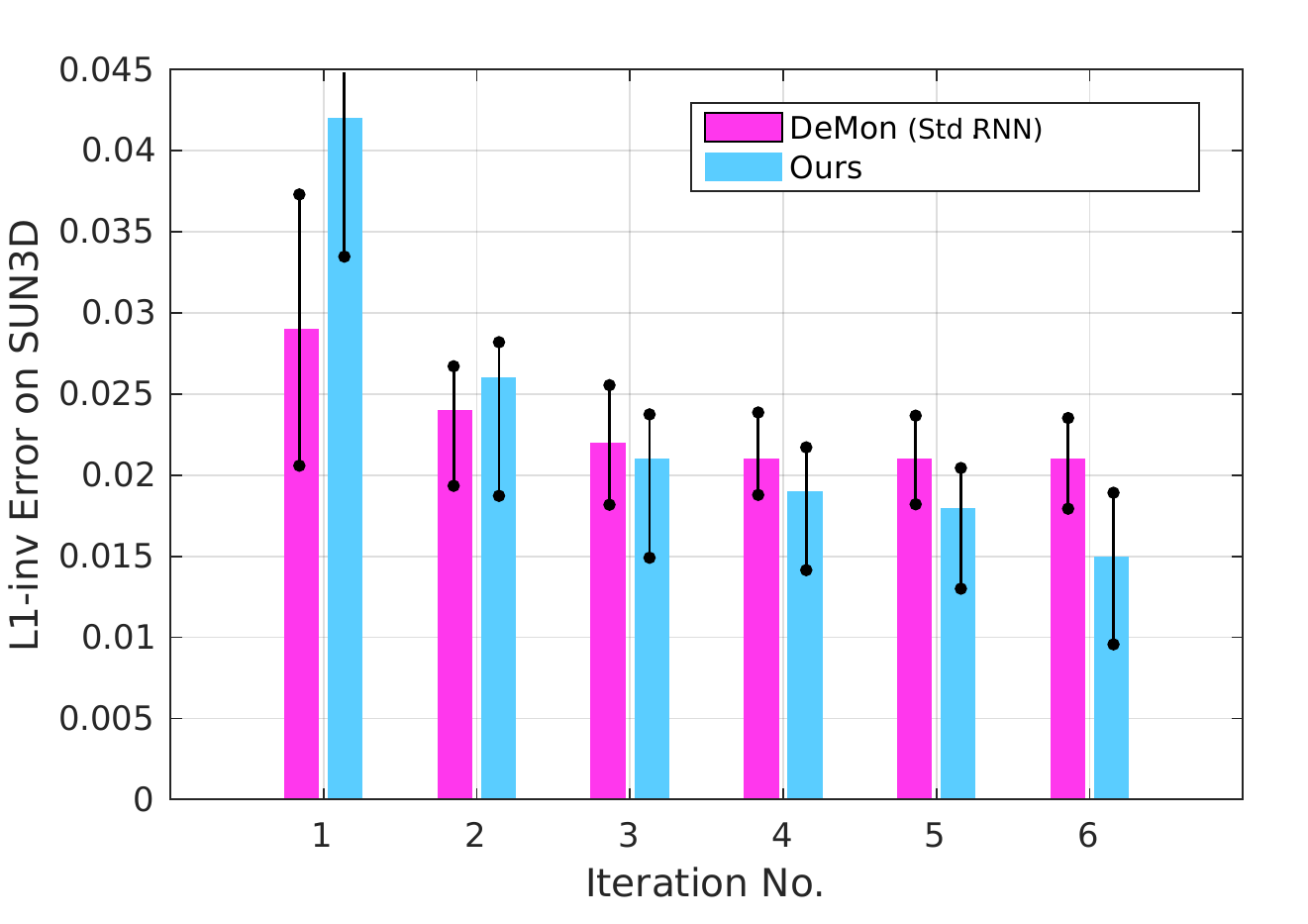}
    \caption{Comparison between our learned optimizer and the (larger) RNN refinement network from DeMon.}
    \label{fig:ablation_graph}
\end{figure}

We conduct an experiment to verify the efficacy of the learned optimization procedure. The first part of our ablation study considers the effect of increasing the number of optimization iterations. These results are shown in Table \ref{tbl:ablation_1} and a qualitative overview of the operation of our network is shown in Figure \ref{fig:results} which visualizes the learned optimization process. The second part of our ablation study evaluates the efficacy of the learned optimizer compared to DeMoN's iterative network. This is show in Figure \ref{fig:ablation_graph}.

\begin{table}[h!]
\centering
{
  \begin{tabular}{c|c|ccc|cc}
\hline

  &             & \multicolumn{3}{c|}{Depth}             & \multicolumn{2}{c}{Motion}    \\
      \hline
& Method      &  L1-inv      &  sc-inv    &  L1-rel    & Rotation        & Translation             \\
      \hline
      \parbox[t]{2mm}{\multirow{4}{*}{\rotatebox[origin=c]{90}{RGB-D}}}  & Initialization  &     0.260        &     0.360      &     0.315      &     2.290      &    27.40              \\
& Opt (5 steps)     &     0.220        &    0.15      &     0.308      &     2.11      &    25.63              \\
& Opt (10 steps) &     0.21        &     0.12      &     0.310      &     1.23      &    24.91 \\
& Opt (15 steps) &     0.019        &     0.09      &     0.301      &     1.01      &    22.14 \\
      \hline

    \end{tabular}
    }
    \vspace{5mm}
    \caption{Results of the ablation study to evaluate the performance of the optimization iterations.}
    \label{tbl:ablation_1}
\end{table}

\subsection{Number of parameters and inference speed}
An advantage of our approach is its parameter efficiency. Compared to DeMoN, our model has significantly fewer parameters. The DeMoN network contains 45,753,883 wheres ours has only  11,438,470 -- making it over $3\times$ more parameter efficient. Ours also has an advantage in terms of inference speed, as although we have to compute the large Jacobian, it still runs around $1.5\times$ faster during inference compared to DeMoN.

\section{Conclusion}

In this paper we have presented an approach for robustly solving nonlinear least squares optimization problems by integrating deep neural models with traditional knowledge of the optimization structure. Our method is based on a novel nonlinear least squares optimizer which is trained to robustly optimize the residuals. Although it is generally applicable to any least squares problem, we have demonstrated the proposed method on the real-world problem of computing depth and egomotion for frames of a monocular video sequence.  Our method can cope with images captured from a wide baseline. In future work we plan to investigate means of increasing the number of residuals that are optimized and thereby achieve an even more detailed prediction. We also plan to further study the interplay between the recurrent neural network and optimization structure and want to investigate the use of predicted confidence estimates in the learned optimization.


\bibliographystyle{splncs04}

\end{document}